\documentclass[letterpaper, 10 pt, conference]{ieeeconf}  % Comment this line out if you need a4paper

\IEEEoverridecommandlockouts                              % This command is only needed if 
\usepackage{xcolor}
\usepackage[linesnumbered,ruled,vlined]{algorithm2e}
\usepackage{graphicx}
\usepackage{multirow}
\usepackage{amsmath}
\usepackage{color}
\usepackage{amssymb}
\usepackage{booktabs}
\usepackage[colorlinks, linkcolor=red]{hyperref}
\usepackage{array}
\usepackage{float}
\SetKwInput{KwInput}{Input}                % Set the Input
\SetKwInput{KwOutput}{Output} 
\makeatletter
\newcommand\figcaption{\def\@captype{figure}\caption}
\newcommand\tabcaption{\def\@captype{table}\caption}
\makeatother
\newcolumntype{C}[1]{>{\centering\arraybackslash}p{#1}}

\usepackage{cite}
\title{\LARGE \bf
StereoPose: Category-Level 6D Transparent Object Pose \\Estimation from Stereo Images via Back-View NOCS
}

\author{Kai Chen$^{1}$, Stephen James$^{2,3}$, Congying Sui$^{4}$, Yun-Hui Liu$^{1}$, Pieter Abbeel$^{2}$ and Qi Dou$^{1}$% <-this % stops a space
\thanks{This work was supported by the Hong Kong Centre for Logistics Robotics. $^{1}$ K. Chen, Y. H. Liu and Q. Dou are with The Chinese University of Hong Kong. $^{2}$ S. James and P. Abbeel are with University of California, Berkeley. $^{3}$ S. James is also with Dyson Robot Learning Lab. $^{4}$ C. Sui is with Hong Kong Centre for Logistics Robotics.}
}

\begin{document}

\maketitle
\thispagestyle{empty}
\pagestyle{empty}

%%%%%%%%%%%%%%%%%%%%%%%%%%%%%%%%%%%%%%%%%%%%%%%%%%%%%%%%%%%%%%%%%%%%%%%%%%%%%%%%
\begin{abstract}
Most existing methods for category-level pose estimation rely on object point clouds. However, when considering transparent objects, depth cameras are usually not able to capture meaningful data, resulting in point clouds with severe artifacts.
Without a high-quality point cloud, existing methods are not applicable to challenging transparent objects. 
To tackle this problem, we present StereoPose, a novel stereo image framework for category-level object pose estimation, ideally suited for transparent objects. 
For a robust estimation from pure stereo images, we develop a pipeline that decouples category-level pose estimation into object size estimation, initial pose estimation, and pose refinement. 
StereoPose then estimates object pose based on representation in the normalized object coordinate space~(NOCS).
To address the issue of image content aliasing, we further define a back-view NOCS map for the transparent object. 
The back-view NOCS aims to reduce the network learning ambiguity caused by content aliasing, and leverage informative cues on the back of the transparent object for more accurate pose estimation.
To further improve the performance of the stereo framework, StereoPose is equipped with a parallax attention module for stereo feature fusion and an epipolar loss for improving the stereo-view consistency of network predictions.
Extensive experiments on the public TOD dataset demonstrate the superiority of the proposed StereoPose framework for category-level 6D transparent object pose estimation.
Project homepage: \url{www.cse.cuhk.edu.hk/~kaichen/stereopose.html}.
\end{abstract}

\section{Introduction}\label{sec:introduction}

% (pose is important)
Transparent objects are common in both daily life and industry. Parsing their poses before grasping or manipulating them using robots is of great importance for ensuring accuracy and safety.
% Since these objects are usually very fragile, parsing their poses before grasping or manipulating them is of great importance for safe robot automation.
% (pose is challenging)
% (how to from object pose estimation to category-level object pose estimation?)
Because of the generalization ability to novel objects, category-level object pose estimation~\cite{wang2019normalized,deng2022icaps,liu2022catre,irshad2022shapo} has received increasing attention recently. 
However, this problem is especially challenging for transparent objects.

\begin{figure}
    \centering
    \begin{minipage}{1.0\linewidth}
        \begin{minipage}{0.24\linewidth}
            \includegraphics[width=1.0\linewidth]{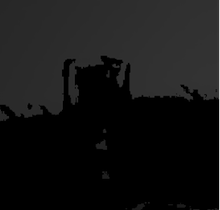}
            \centering{(a)}
        \end{minipage}
        \hfill
        \begin{minipage}{0.24\linewidth}
            \includegraphics[width=1.0\linewidth]{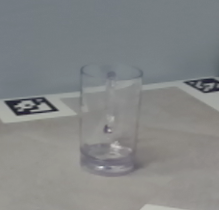}
            \centering{(b)}
        \end{minipage}
        \hfill
        \begin{minipage}{0.24\linewidth}
            \includegraphics[width=1.0\linewidth]{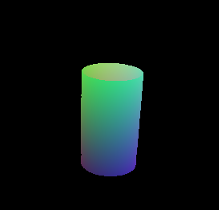}
            \centering{(c)}
        \end{minipage}
        \hfill
        \begin{minipage}{0.24\linewidth}
            \includegraphics[width=1.0\linewidth]{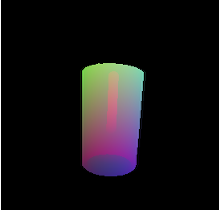}
            \centering{(d)}
        \end{minipage}
    \end{minipage}
    \vfill
    \vspace{2mm}
    \begin{minipage}{1.0\linewidth}
        \includegraphics[width=1.0\linewidth]{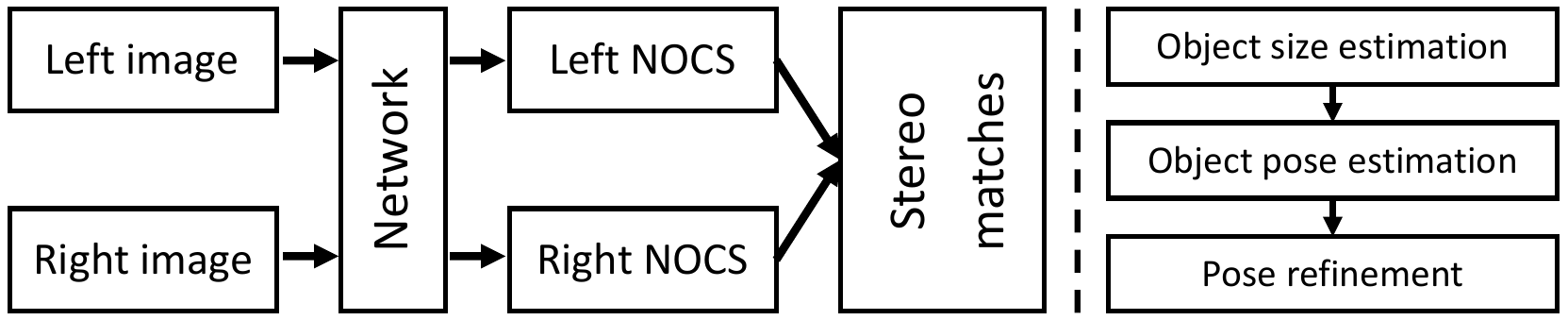}
    \end{minipage}
    \vspace{-1mm}
    \caption{Challenges of category-level transparent object pose estimation and an overview of the proposed StereoPose framework. (a) The depth map of a transparent mug with severe artifacts. (b) The RGB image patch of a transparent mug with content aliasing. (c) The conventional front-view NOCS map. (d) The proposed back-view NOCS map.}\vspace{-6mm}
    \label{fig:cover_page}
\end{figure}

First of all, transparent objects usually have specular and reflective materials. 
These factors make acquiring high-quality depth maps of transparent objects challenging with commonly used depth sensors. 
As shown in Fig.~\ref{fig:cover_page}~(a), their depth maps often exhibit severe artifacts, such as noise and large missing areas.
Unfortunately, existing state-of-the-art methods~\cite{chen2021sgpa,liu2022catre} rely on high-quality depth maps and point clouds to handle intra-class shape variation of different objects for accurate object pose estimation.
% require explicitly using the point cloud recovered from the depth map to handle intra-class shape variation of different objects. 
% They highly rely on a high-quality depth map for accurate category-level object pose estimation. 
The point cloud artifacts caused by the transparent material are hard to mitigate even with recent techniques such as depth completion and reconstruction~\cite{sajjan2020clear,zhu2021rgb,xu2021seeing,fang2022transcg}. Without a means to guarantee point cloud quality, these point-cloud-based pose estimation methods perform poorly on transparent objects.

% There are few devices that are able to capture high-quality depth maps for transparent objects.
% Unfortunately, existing state-of-the-art methods~\cite{chen2021sgpa,liu2022catre} require explicitly using the point cloud recovered from the depth map to handle intra-class shape variation of different objects. They highly rely on a high-quality depth map for accurate category-level object pose estimation.  Without a good way to guarantee the point cloud quality, these point-cloud-based methods are not applicable to transparent objects.

While depth data fail to fully capture transparent objects, RGB data can often exhibits severe content aliasing caused by the transparent material. 
As shown in Fig.~\ref{fig:cover_page}~(b), the handle of the mug could be observed on the image, despite being on the back of the mug. 
The mixed image content brings challenges to predicting dense correspondences of image pixels for object pose estimation.
It is because conventional coordinate representations~(e.g., instance-level voting field~\cite{peng2019pvnet} and category-level NOCS map~\cite{wang2019normalized}) used on non-transparent objects only describe dense correspondences for the front-view\footnote{Front view denotes the side of an object facing the camera.} of the object.
As shown in Fig.~\ref{fig:cover_page}~(b) and (c), the RGB image and the conventional front-view NOCS map are highly inconsistent in appearance.
Learning this NOCS map from the aliased image has a relatively large ambiguity, which decreases the quality of the predicted NOCS map and makes the category-level pose estimation results not accurate. 

% An alternative solution is predicting a NOCS map~(a.k.a, a front-view NOCS map) purely based on the image feature~\cite{tian2020shape}. The NOCS map is a canonical representation~\cite{wang2019normalized} that provides dense correspondences between the camera frame and the object frame for recovering the category-level object pose information.
% However, RGB images of transparent objects often exhibit severe content aliasing caused by the transparent material. As shown in Fig.~\ref{fig:cover_page}, the \textit{`handle'} of the mug can be observed on the image, despite being on the back of the mug.
% %  which is very different from the image of a non-transparent mug.
% The image content aliasing increases the ambiguity and difficulty of estimating the front-view NOCS map from the image, as they are not visually consistent~(compare Fig.~\ref{fig:cover_page}~(a) and Fig.~\ref{fig:cover_page}~(c)). 
% It brings additional challenges for category-level transparent object pose estimation based on images. 

% On the one hand, the aliasing increases the ambiguity of object pose learning from RGB images. On the other hand, the aliasing inspires us to think about one question: is it possible to leverage the information on the back of the transparent object for more accurate object pose estimation?

To address the above issues, we present StereoPose, a novel stereo image framework for category-level 6D transparent object pose estimation.
Instead of explicitly using the object point cloud, StereoPose exploits stereo images to implicitly model the object shape information, which is important for category-level object pose estimation. 
As shown in Fig.~\ref{fig:cover_page}, StereoPose estimates NOCS maps for each view of the stereo images.
An individual NOCS map provides intra-view correspondences between the camera and the object frame, while the left and right NOCS maps provide inter-view correspondences for the stereo views. 
Based on these correspondences, StereoPose decouples the estimation of category-level object pose into object size estimation, initial pose estimation, and pose refinement, which achieves a robust object pose estimation without using the depth map.

Due to the transparent property, the back side of the object can always be observed from the image. 
Different from opaque object pose estimation, some object structures on the back~(e.g., bottle rim or mug handle) can also be informative cues for determining object pose but are neglected in existing methods.
To tackle this issue,
% In order to tackle the problem of image content aliasing, 
as shown in Fig.~\ref{fig:cover_page}~(d), we define a back-view NOCS map for the transparent object. 
Given an aliased image, we jointly estimate the conventional front-view NOCS map and the back-view NOCS map from the image. 
Jointly estimating two NOCS maps reduces the learning ambiguity caused by the image content aliasing, which helps to produce a NOCS map more accurately.
Moreover, the back-view NOCS map captures informative cues on the back of the object, which also reduces the pose estimation ambiguity caused by object self-occlusion and further improves the pose accuracy for transparent objects. 
To make full use of the stereo features, we further propose a parallax attention module to fuse stereo image features through the guidance of parallax information. An epipolar geometry loss is applied to constrain the consistency between NOCS maps of left and right views.
We summarize our main contributions as follows:
\begin{itemize}
    \item We propose StereoPose, a novel stereo image framework for category-level transparent object pose estimation. A parallax attention module is proposed to fuse stereo image features. An elaborated pipeline is designed for robust object pose estimation with pure stereo images.
    \item We define the back-view NOCS map for the transparent objects. It reduces the negative effect of image content aliasing on transparent object pose estimation. It also captures informative structures on the back of the object for more accurate object pose estimation.
    \item We conduct extensive experiments on the public TOD dataset, which covers three typical categories of transparent objects with about 36K stereo images. StereoPose achieves dramatic performance improvements over other existing methods for category-level 6D transparent object pose estimation.
\end{itemize}
\section{Related Works}\label{sec:related_works}
% \subsection{Transparent Object Pose Estimation}
\subsection{Transparent Object Pose Estimation}
Transparent objects~\cite{chen2022clearpose} often exhibit sparkling texture, refractive and reflective material~\cite{ichnowski2021dex}. These factors result in a lot of artifacts on their image appearance and point cloud geometry, which makes accurate pose estimation for transparent objects very challenging~\cite{zhu2021transfusion}. 
Early methods extract edge information from images for transparent object pose estimation. Phillips et al.~\cite{phillips2016seeing} develop a heuristic method to analyze the shape of the extracted edge curves for pose estimation of rotationally symmetric transparent objects. 
Lysenkov et al.~\cite{lysenkov2013recognition,lysenkov2013pose} extract object silhouettes from the edge information. They assume that the CAD model of the transparent object is available and then estimate object pose by adjusting the pose hypothesis to fit the silhouettes. 
However, the 2D edge information is ambiguous in determining a complete 6D object pose, which limits the pose accuracy. In order to use the 3D geometry feature for more accurate pose estimation, recent methods~\cite{tang2021depthgrasp,fang2022transcg,dai2022domain,jiang2022a4t} do much effort to recover accurate geometry of transparent objects from their defective RGB-D data. 
Sajjan et al.~\cite{sajjan2020clear} present a multi-task learning network to infer surface normal and object boundary of transparent objects for depth refinement. 
Zhu et al.~\cite{zhu2021rgb} complete the missing depth with an implicit neural representation. 
Ichnowski et al.~\cite{ichnowski2021dex} use neural radiance fields to estimate the geometry of transparent objects. 
Xu et al.~\cite{xu20206dof} develop a two-stage network to jointly refine the depth map and perform pose estimation for transparent objects.
However, training an additional network for depth refinement is costly. The refined depth cannot recover the object shape perfectly, and the residual shape distortion would affect the final pose accuracy.
In contrast, our method achieves accurate object pose estimation with stereo images.

\subsection{Category-Level Object Pose Estimation}
Category-level object pose estimation aims to predict the pose of novel objects. 
The object intra-class shape variation~\cite{chen2020learning,li2021leveraging} is the main challenge of applying the network to novel objects for accurate pose estimation.
In this regard, previous approaches~\cite{sahin2018category,sahin2019instance} extract shape-invariant features from depth map for category-level pose estimation. 
Wang et al.~\cite{wang2019normalized} then present the NOCS representation, which represents objects in a normalized canonical space to reduce the shape difference for object pose estimation. 
% reduce object shape difference by representing objects in the NOCS space. Dense correspondences are estimated for pose estimation by predicting a NOCS map from the RGB-D observation of the object.
Furthermore, Tian et al.~\cite{tian2020shape} describe a prior-based framework to explicitly reconstruct the novel object shape in the NOCS space. 
This prior-based framework is widely applied to recent methods with further improvements, such as recurrent NOCS model reconstruction~\cite{wang2021category}, dynamic prior adaptation~\cite{chen2021sgpa}, and deep prior deformation~\cite{lin2022category}.
Although the NOCS representation and the prior-based framework are quite popular and have achieved state-of-the-art performance for category-level object pose estimation, nearly all existing methods~\cite{irshad2022centersnap,di2022gpv,lin2022sar,zhang2022rbp} highly rely on point cloud data and are therefore not suited for transparent object category-level pose estimation.
Our method is a complement to existing methods and is tailored for transparent object category-level pose estimation.
\section{Methodology}\label{sec:method}
In this section, we present the proposed StereoPose for category-level transparent object pose estimation. In Sec.~\ref{subsec:framework}, we present the overall framework based on stereo images. A set of steps are elaborated for accurate pose estimation from pure stereo images. In Sec.~\ref{subsec:back_view_nocs}, we introduce the back-view NOCS map for transparent objects. We propose to jointly estimate the front-view and back-view NOCS maps to address the problem of image content aliasing for transparent object pose estimation. In Sec.~\ref{subsec:stereo_fusion}, we present a parallax attention module to integrate features from stereo images for more accurate object pose estimation with complementary features of stereo images.
\begin{figure*}
    \centering
    \begin{minipage}{0.9\linewidth}
    \begin{minipage}{1.0\linewidth}
        \begin{minipage}{0.30\linewidth}
            \includegraphics[width=1.0\linewidth]{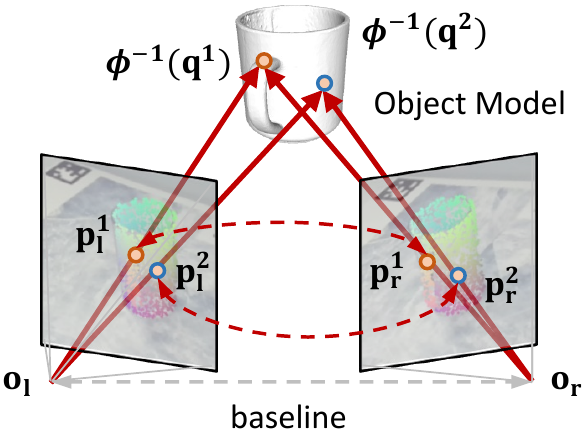}
        \end{minipage}
        \hfill
        \begin{minipage}{0.34\linewidth}
            \includegraphics[width=1.0\linewidth]{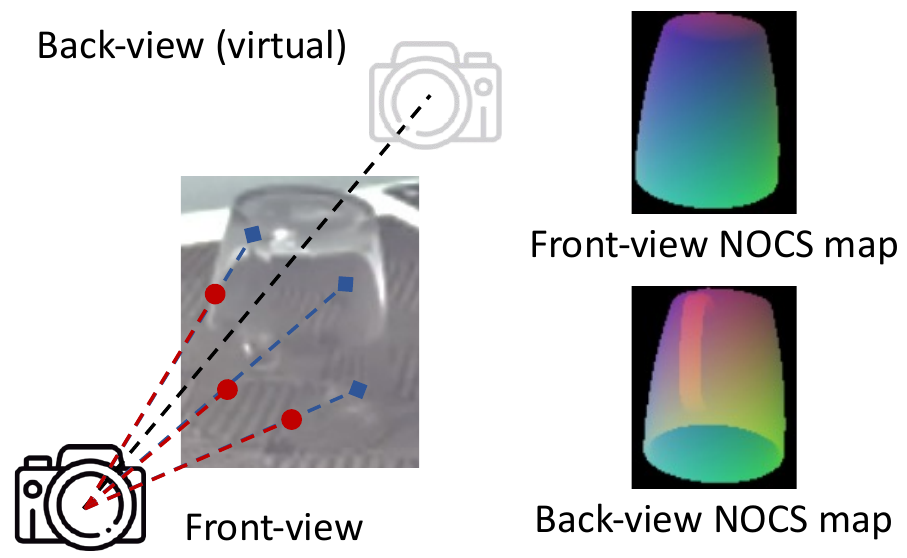}
        \end{minipage}
        \hfill
        \begin{minipage}{0.30\linewidth}
            \includegraphics[width=1.0\linewidth]{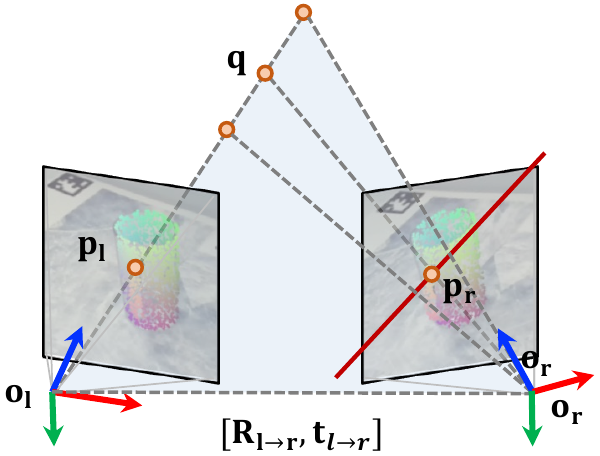}
        \end{minipage}
    \end{minipage}
    \vfill
    \vspace{1mm}
    \begin{minipage}{1.0\linewidth}
        \begin{minipage}{0.30\linewidth}
            \centering{(a)}
        \end{minipage}
        \hfill
        \begin{minipage}{0.34\linewidth}

            \centering{(b)}
        \end{minipage}
        \hfill
        \begin{minipage}{0.30\linewidth}
            \centering{(c)}
        \end{minipage}
    \end{minipage}
    \vspace{-7mm}
    \caption{(a) Finding cross-view correspondences based on the predicted NOCS coordinates. (b) An illustration of generating the back-view NOCS map. (c) Epipolar geometry loss for constraining the consistency of left-view and right-view NOCS maps.}\vspace{-6mm}
    \end{minipage}
    \label{fig:method}
\end{figure*}

\subsection{Overview of StereoPose}\label{subsec:framework}
Different from existing point-cloud-based methods, our proposed StereoPose leverages stereo image features of the transparent object to account for the inherent shape variation of novel objects and predicts NOCS maps to construct dense correspondences for category-level object pose estimation. Following~\cite{tian2020shape}, StereoPose is developed based on a similar prior-based framework. Let $\mathbf{F}_l,\mathbf{F}_r\in\mathbb{R}^{H\times W\times C}$ be stereo features extracted from stereo images, $\mathcal{P}\in\mathbb{R}^{N\times3}$ be the average prior model for the corresponding object category. StereoPose predicts a deformation field $\mathcal{D}\in\mathbb{R}^{N\times3}$ to reconstruct the shape of the observed transparent object by $\mathcal{P}'=\mathcal{P}+\mathcal{D}$. Subsequently, StereoPose predicts two matching matrices $\mathcal{A}_l, \mathcal{A}_r\in\mathcal{R}^{M\times N}$ to predict NOCS coordinates for $M$ sampled image pixels of the target transparent object as $\mathcal{M}_{(l,r)}=\mathcal{A}_{(l,r)}\times \mathcal{P}'$.

For one sampled image pixel $\mathbf{p}\in\mathbb{R}^2$, its NOCS coordinates on the corresponding NOCS map could be denoted as $\mathbf{q}\in\mathbb{R}^3$. According to the definition of NOCS in~\cite{wang2019normalized}, $\mathbf{p}$ and $\mathbf{q}$ should satisfy:
\begin{equation}\label{eq:nocs}
    \dbinom{\mathbf{p}}{1}=\frac{1}{Z}\mathbf{K}
    \begin{bmatrix}
        s\mathbf{R} & \mathbf{t}\\
        \mathbf{0^{\top}} & 1
    \end{bmatrix}\dbinom{\mathbf{q}}{1},
\end{equation}
where $\mathbf{K}$ denotes the camera intrinsic matrix, $s$ denotes the isotropic scale factor related to the object size, and $[\mathbf{R, t}]$ denotes the transformation matrix corresponding to the 6D object pose. Given a calibrated camera with known intrinsic parameters, category-level object pose estimation aims to recover $s$ and $[\mathbf{R, t}]$ simultaneously. Conventional methods first recover the object point cloud in the camera frame using a depth map, and then jointly estimate $s, \mathbf{R}$ and $\mathbf{t}$ by 3D-3D registration. However, it is not applicable to transparent objects with low-quality depth maps.

StereoPose therefore proposes to decouple the estimation of $s$ and $\mathbf{[R, t]}$. An elaborated pipeline then is developed for a robust and accurate category-level pose estimation of transparent objects with pure stereo images. Firstly, we estimate $s$ based on correspondences between $\mathcal{M}_l$ and $\mathcal{M}_r$. As shown in Fig.~\ref{fig:method}~(a), we propose to efficiently find the cross-view correspondences based on the predicted NOCS coordinates. Let $\mathbf{p}_l, \mathbf{p}_r\in \mathbb{R}^2$ be two points on the left and right image. Intuitively, $\mathbf{p}_l$ and $\mathbf{p}_r$ are a pair of correspondences if they have the same NOCS coordinates. For a stereo camera with a known baseline length $b$, we can measure the depth value of the corresponding 3D point as $d=\frac{f}{\|\mathbf{p}_l-\mathbf{p}_r\|}\times b$. For arbitrary two pairs of points, we can estimate one scale factor based on the ratio of the distance in the camera frame to the distance in the NOCS space. For a robust estimation, we extract all matched points from $\mathcal{M}_l$ and $\mathcal{M}_r$ and average the estimation results across all pairs.

Given the estimated $s$, StereoPose further solves the remaining $\mathbf{R}$ and $\mathbf{t}$ by the Perspective-n-Point~(PnP) algorithm~\cite{lepetit2009epnp}. Our method directly leverages 2D-3D correspondences for object pose estimation. Using the RANSAC algorithm~\cite{fischler1981random} would further improve the robustness of the estimation results. We also compare our proposed decoupled pose estimation pipeline with the conventional scheme, in which we first recover 3D positions for all matched points and then jointly estimate $s$, $\mathbf{R}$ and $\mathbf{t}$ by fitting a 3D-3D similarity transformation. Results indicate that our proposed method could achieve higher performance for category-level transparent object pose estimation. To further improve the pose accuracy, similar to~\cite{haugaard2022surfemb}, we propose to rescale $\mathbf{t}$, which aims to align the average object depth corresponding to the estimated object pose to the average depth value recovered from matched points of $\mathcal{M}_l$ and $\mathcal{M}_r$.

\subsection{Back-View NOCS Map}\label{subsec:back_view_nocs}
Though our proposed StereoPose manages to estimate pose for transparent objects from pure stereo images with an elaborated pose estimation pipeline, the problem of image content aliasing has not been addressed for transparent objects. This unresolved problem would significantly affect the pose accuracy. It is because (1) Estimating a conventional front-view NOCS map from the aliased image is of a large ambiguity, which affects the accuracy of the predicted NOCS map; (2) Though the back side could be observed on the RGB image of a transparent object, the conventional front-view NOCS map fails to leverage this information to improve the pose estimation accuracy.

In order to tackle these problems, we define the back-view NOCS map for transparent objects. Fig.~\ref{fig:method}~(b) illustrates the generation process. Let $\mathbf{o}\in\mathbb{R}^3$ be the camera center, and $\mathbf{r}$ be a ray marching from $\mathbf{o}$. Let $\mathbf{C}$ be the point set that $\mathbf{r}$ intersects with the object model. The conventional front-view NOCS could be denoted as:\vspace{-1mm}
\begin{equation}
    \mathbf{c}_{front}=\phi(\arg\min_{\mathbf{c}\in\mathbf{C}}\|\mathbf{c}-\mathbf{o}\|),\vspace{-1mm}
\end{equation}
where $\phi(\cdot)$ returns the NOCS coordinates for the corresponding point on the object model. For the back view, instead of really moving the camera to the back side for capturing a back-view NOCS map, we take the NOCS coordinates of the object point along $\mathbf{r}$ that is the farthest from the front-view camera as the corresponding back-view NOCS coordinates. Specifically, the back-view NOCS map is defined as:
\begin{equation}
    \mathbf{c}_{back}=\phi(\arg\max_{\mathbf{c}\in\mathbf{C}}\|\mathbf{c}-\mathbf{o}\|).
\end{equation}

According to our definition, the back-view NOCS map would be ideally aligned with the front-view NOCS map as well as the RGB image. It facilitates estimating the back-view NOCS map from the RGB image. Fig.~\ref{fig:method}~(b) presents an example for the generated back-view NOCS map. Our proposed back-view NOCS map exhibits the informative structure cues on the back of the object. We then distill the front-view feature and the back-view feature from the aliased image to jointly estimate the front-view and the back-view NOCS maps. It decreases the ambiguity of network learning and helps to improve the NOCS map accuracy. In addition, the proposed StereoPose adopts the PnP solver for object pose estimation. It is based on 2D-3D correspondences, which means that the predicted back-view NOCS map could be directly applied to the PnP scheme for object pose estimation. In other words, the introduction of the back-view NOCS map excavates the structure information on the back of the transparent object for more accurate pose estimation.

\subsection{Parallax Attention for Stereo Feature Fusion}\label{subsec:stereo_fusion}
Stereo images provide informative cross-view features for the target object. Properly fusing them enhances the image feature for representing the shape of the transparent object for more accurate category-level object pose estimation. The parallax\footnote{Parallax is a position displacement of matched points on stereo images.} magnitude of an object exhibited on stereo images would vary with the shape and depth of the object. In other words, the object parallax implicitly indicates this useful information, which could be used to guide the stereo feature fusion for improving the object pose estimation performance.

In this regard, we propose a parallax attention module for stereo feature fusion. Given the stereo image features $\mathbf{F}_l,\mathbf{F}_r\in\mathbb{R}^{H\times W\times C}$, we compute the pixel-wise feature correlation between $\mathbf{F}_l$ and $\mathbf{F}_r$ to softly models the parallax information of an object with respect to the camera. We refer to this pixel-wise correlation as a parallax attention map and use it for stereo feature fusion. Specifically, we leverage the disentangled non-local block~\cite{yin2020disentangled,wang2021symmetric} to compute the parallax attention map. At first, two whiten layers are used to normalize $\mathbf{F}_l$ and $\mathbf{F}_r$ to $\mathbf{F'}_l$ and $\mathbf{F'}_r$. After that, the attention map could be computed as:\vspace{-2mm}
\begin{equation}
    A_{r\rightarrow l}=\sigma(\mathbf{F'}_l\otimes \mathbf{F'}_r^\top), A_{l\rightarrow r}=\sigma(\mathbf{F'}_r\otimes\mathbf{F'}_l^\top),\vspace{-2mm}
\end{equation}
where $\otimes$ denotes the batch-wise matrix multiplication, $\sigma(\cdot)$ is the operation of softmax normalization along the last dimension. The resulted $A_{r\rightarrow l}, A_{l\rightarrow r}\in\mathbb{R}^{H\times W\times W}$ are parallax attention maps from the right view to the left view and from the left view to the right view, respectively. To fuse the right-view feature with the left-view one, we convert $\mathbf{F}_r$ to the left-view based on $A_{r\rightarrow l}$:\vspace{-2mm}
\begin{equation}
    \mathbf{F}_{r\rightarrow l}=A_{r\rightarrow l}\otimes \mathbf{F}_r.\vspace{-2mm}
\end{equation}
We can follow a similar way to obtain $\mathbf{F}_{l\rightarrow r}$. The converted features are fused with the original feature by an MLP layer for subsequent NOCS map prediction.

\subsection{Epipolar Geometry Loss Function}\label{subsec:loss}
Since the proposed StereoPose would separately predict NOCS maps for the left view and the right view of the stereo camera, to constrain the inter-view consistency of left-view and right-view NOCS maps, we propose an epipolar geometry loss in this section. Let $\mathbf{K}_l$ and $\mathbf{K}_r$ be the intrinsic matrices of the left camera and the right camera respectively. Let $\mathbf{R}_{l\rightarrow r}$ and $\mathbf{t}_{l\rightarrow r}$ be the relative rotation and translation from the left camera to the right camera, $\mathcal{M}_l$ and $\mathcal{M}_r$ be the left-view NOCS map and the right-view NOCS map. As shown in Fig.~\ref{fig:method}~(c), based on the epipolar geometry~\cite{hartley2003multiple}, if we follow the scheme mentioned in Sec.\ref{subsec:framework} to find a pair of matched points $(\mathbf{p}_l, \mathbf{p}_r)$ that have the same coordinates in the NOCS space, $\mathbf{p}_l$ and $\mathbf{p}_r$ should satisfy:
\begin{equation}
    \hat{\mathbf{p}}_l\mathbf{K}_l^{-\top}[\mathbf{t}_{l\rightarrow r}]_{\times}\mathbf{R}_{l\rightarrow r}\mathbf{K}_r^{-1}\hat{\mathbf{p}}_r^{\top}=0,
\end{equation}
where $\hat{\mathbf{p}}_l,\hat{\mathbf{p}}_r \in \mathbb{R}^{1\times3}$ are homogeneous image pixel coordinates, $[\cdot]_{\times}$ denotes computing the skew symmetric matrix of $\mathbf{t}_{l\rightarrow r}$. The epipolar geometry loss $l_{ep}$ therefore could be directly defined as $l_{ep}=\hat{\mathbf{p}}_l\mathbf{K}_l^{-\top}[\mathbf{t}_{l\rightarrow r}]_{\times}\mathbf{R}_{l\rightarrow r}\mathbf{K}_r^{-1}\hat{\mathbf{p}}_r^{\top}$.

To train the StereoPose model, apart from $l_{ep}$, similar to~\cite{tian2020shape,wang2021category}, we also adopt a reconstruction loss $l_{cd}$ to compute the Chamfer Distance between $\mathcal{P}'$ and $\mathcal{P}_{gt}$ as:
\begin{equation}
    l_{cd}=\sum_{x\in\mathcal{P}'}\min_{y\in\mathcal{P}_{gt}}\|x-y\|^2_2 + \sum_{y\in\mathcal{P}_{gt}}\min_{x\in\mathcal{P}'}\|x-y\|^2_2.
\end{equation}
A NOCS loss $l_{nocs}$ is used to minimize the $L_1$ distance between $\mathcal{M}$ and $\mathcal{M}_{gt}$.
In addition, we regularize the deformation field with $l_d=\frac{1}{N}\sum_{\mathbf{d}\in\mathcal{D}}\|\mathbf{b}\|_2$, and constrain the distribution of $\mathcal{A}$ by $l_a=\frac{1}{M}\sum_{i}\sum_{j}-a_{i,j}\;\text{log}\ a_{i,j}$.
In summary, the total loss function for training StereoPose is:
\begin{equation}
    L=\lambda_1l_{ep} + \lambda_2l_{cd} + \lambda_3l_{nocs} + \lambda_4l_a + \lambda_5l_d.
\end{equation}
\section{Experiments}\label{sec:experiment}
\subsection{Datasets and Evaluation Metrics}\label{subsec:dataset}
We evaluate our proposed StereoPose on the TOD dataset~\cite{liu2020keypose}. This dataset contains three categories of transparent objects: bottle~(3 instances), mug~(7 instances), and cup~(2 instances), with about 36K stereo image pairs for 12 different object instances in 10 different environment backgrounds.  We follow the setting in~\cite{liu2020keypose} to conduct category-level experiments across three different category splits: `mug', `bottle', and `bottle and cup'. In each split, we train StereoPose on different instances and evaluate the model on novel instances that are not used during training. 

\begin{table*}[thb]
    \centering
    \begin{minipage}{1.0\linewidth}
        \caption{Quantitative comparison with state-of-the-art methods on TOD Dataset. The larger the metric, the higher the pose accuracy.}\vspace{-3mm}
    \label{table:sota_comparison}
    \resizebox{\textwidth}{!}{%
    \renewcommand{\arraystretch}{1.2}
    \begin{tabular}{l|cccc|cccc|cccc}
        \hline
        \multirow{2}{*}{Method} & \multicolumn{4}{c|}{Bottle} & \multicolumn{4}{c|}{Bottle and Cup} & \multicolumn{4}{c}{Mug} \\\cline{2-13}
        % \midrule
        {} & $3D_{25}$ & $3D_{50}$ & $10^{\circ}5cm$ & $10^{\circ}10cm$ & $3D_{25}$ & $3D_{50}$ & $10^{\circ}5cm$ & $10^{\circ}10cm$ & $3D_{25}$ & $3D_{50}$ & $10^{\circ}5cm$ & $10^{\circ}10cm$ \\\hline
        SPD~\cite{tian2020shape} & 44.3 & 7.4 & 11.5 & 17.8 & 41.8 & 6.5 & 7.8 & 12.2 & 63.6 & 19.7 & 2.3 & 4.2 \\
        SGPA~\cite{chen2021sgpa} & 46.9 & 9.6 & 13.3 & 22.5 & 48.1 & 12.5 & 9.5 & 14.6 & 64.4 & 19.6 & 2.8 & 5.1 \\
        KeyPose~\cite{liu2020keypose} & - & - & 52.7 & 62.3 & - & - & 30.2 & 35.1 & - & - & 24.6 & 25.1 \\\hline
        StereoPose  & \textbf{85.4} & \textbf{22.2} & \textbf{57.8} & \textbf{70.3} & \textbf{85.7} & \textbf{17.8} & \textbf{34.7} & \textbf{41.3} & \textbf{97.9} & \textbf{77.4} & \textbf{34.4} & \textbf{38.2} \\ \hline
    \end{tabular}
    }\vspace{-6mm}
    \end{minipage}
\end{table*}

We quantitatively evaluate the performance of category-level object pose estimation following the widely used metrics in~\cite{tian2020shape,chen2021sgpa,you2022cppf}. Specifically, the metric of 3D IoU computes the overlapping ratio of 3D bounding boxes corresponding to the ground-truth pose and the predicted pose. We report the mAP with respect to $\text{3D}_{25}$ and $\text{3D}_{50}$ in our experiments. In addition, we directly measure the rotation and translation error between the ground-truth pose and the predicted pose. We report mAP with respect to $10^{\circ}5cm$ and $10^{\circ}10cm$ in our experiments.

\subsection{Implementation Details}\label{subsec:implementation}
StereoPose uses a ResNet-18~\cite{he2016deep} with PSP network~\cite{zhao2017pyramid} to extract features for stereo images. Similar to most category-level object pose estimation methods, the target object region is first cropped based on the segmentation result. We train a TransLab~\cite{xie2020segmenting} model on Trans10K dataset and then directly use the trained model for the transparent object segmentation of the TOD dataset. The cropped object image patch is resized to $224\times224$ and then fed into the backbone network for image feature extraction. We set $M=N=1024$ when reconstructing the NOCS object model and predicting the NOCS map for the observed image under the prior-based framework. For the hyberparameters of the loss function, we empirically use $\lambda_1=0.01, \lambda_2=5.0, \lambda_3=1.0, \lambda_4=0.0001,$ and $\lambda_5=0.01$.

\subsection{Comparison with State-of-the-Art Methods}\label{subsec:sota_comparison}
Since there are few available methods designed for stereo category-level transparent object pose estimation, we adapt existing methods to transparent objects and compare our proposed framework with three state-of-the-art methods. On the one hand, we compare our method with the paradigm of `depth completion + opaque object pose estimation'. We first train TransCG~\cite{fang2022transcg} on the TOD dataset and use it to improve the depth map. After that, we use the refined depth map for category-level object pose estimation. We choose two state-of-the-art methods~(SPD~\cite{tian2020shape} and SGPA~\cite{chen2021sgpa}) for comparisons. On the other hand, we compare our method with one keypoint-based approach~(KeyPose~\cite{liu2020keypose}), in which stereo images are used to predict category-level object key points. These key points then are used to recover the complete object pose. Note that KeyPose does not need depth/point cloud, which thus could be applied to transparent object pose estimation without any modification.

\begin{figure}[t]
	\centering
		\begin{minipage}{1.0\linewidth}
			\centering
            \tabcaption{Evaluation of the depth accuracy before and after TransCG refinement. For both `RMSE' and `MAE', the smaller the value, the higher the accuracy.}
            \vspace{-2mm}
            \label{tab:evaluation_depth}
            \resizebox{1.0\linewidth}{!}{%
            \renewcommand{\arraystretch}{1.0}
            \begin{tabular}{l|cc|cc|cc}
                \hline
               {} & \multicolumn{2}{c|}{Bottle} & \multicolumn{2}{c|}{Cup} & \multicolumn{2}{c}{Mug} \\\cline{2-7}
               {} & RMSE  & MAE  & RMSE  & MAE  & RMSE  & MAE \\\hline
               Original & {0.506} & {0.450} & {0.516} & {0.466} & {0.564} & {0.515}\\
               Refined & {\textbf{0.071}} & {\textbf{0.065}} & {\textbf{0.075}} & {\textbf{0.068}} & {\textbf{0.080}} & {\textbf{0.072}}\\\hline
            \end{tabular}
            }
		\end{minipage}
		\vfill
		\vspace{1mm}
		\begin{minipage}[h]{1.0\linewidth}
			\begin{minipage}{1.0\linewidth}
                \begin{minipage}{0.24\linewidth}
                    \includegraphics[width=1.0\linewidth]{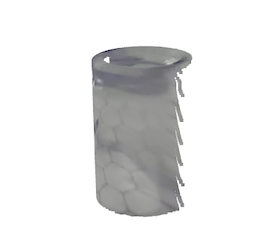}
                \end{minipage}
                \hfill
                \begin{minipage}{0.24\linewidth}
                    \includegraphics[width=1.0\linewidth]{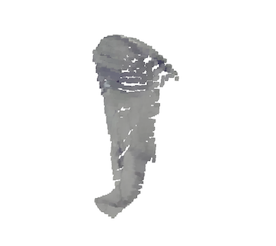}
                \end{minipage}
                \hfill
                \begin{minipage}{0.24\linewidth}
                    \includegraphics[width=1.0\linewidth]{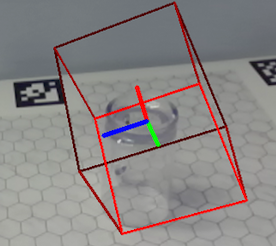}
                \end{minipage}
                \hfill
                \begin{minipage}{0.24\linewidth}
                    \includegraphics[width=1.0\linewidth]{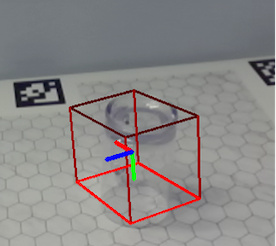}
                \end{minipage}
            \end{minipage}
            \vfill
            \vspace{1mm}
            \begin{minipage}{1.0\linewidth}
                \begin{minipage}{0.24\linewidth}
                    \includegraphics[width=1.0\linewidth]{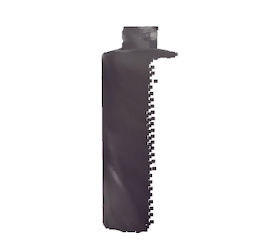}
                    \centering{(a)}
                \end{minipage}
                \hfill
                \begin{minipage}{0.24\linewidth}
                    \includegraphics[width=1.0\linewidth]{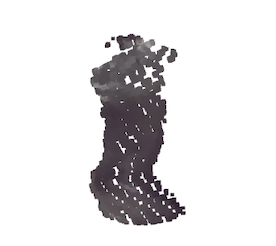}
                    \centering{(b)}
                \end{minipage}
                \hfill
                \begin{minipage}{0.24\linewidth}
                    \includegraphics[width=1.0\linewidth]{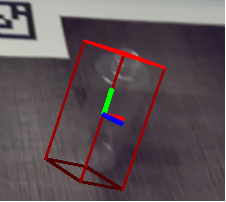}
                    \centering{(c)}
                \end{minipage}
                \hfill
                \begin{minipage}{0.24\linewidth}
                    \includegraphics[width=1.0\linewidth]{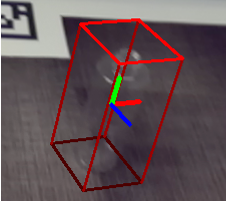}
                    \centering{(d)}
                \end{minipage}
            \end{minipage}\vspace{-3mm}
            \caption{\textbf{Qualitative comparisons with `depth completion + opaque object pose estimation'.} (a) Ground-truth point cloud. (b) Recovered point cloud by TransCG. (c) Results of SGPA. (d) Results of StereoPose.}
            \label{fig:pose_comparison_depth}
		\end{minipage}\vspace{-5mm}
\end{figure}

Tab.~\ref{table:sota_comparison} presents the comparative results. StereoPose significantly outperforms other competing methods in all evaluation metrics. First of all, StereoPose achieves higher accuracy than `depth completion + opaque object pose estimation'. Although the depth accuracy indeed gets apparent improvement after the refinement of TransCG~(as shown in Tab.~\ref{tab:evaluation_depth}), the point cloud recovered from the refined depth map still exhibits severe distortion, which cannot represent the shape of the transparent object precisely. State-of-the-art SPD and SGPA are easily affected by the point cloud artifacts. As shown in Fig.~\ref{fig:pose_comparison_depth}, without a good point cloud to represent the transparent object shape, their pose estimation accuracy is inferior to StereoPose by a large margin. 

\begin{figure*}
    \centering
    \begin{minipage}{0.95\linewidth}
        \begin{minipage}{0.16\linewidth}
            \begin{minipage}{1.0\linewidth}
                \includegraphics[width=1.0\linewidth]{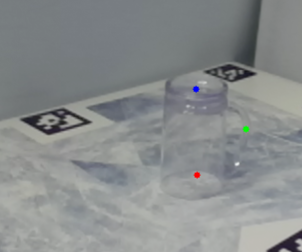}
            \end{minipage}
            \vfill
            \vspace{1mm}
            \begin{minipage}{1.0\linewidth}
                \includegraphics[width=1.0\linewidth]{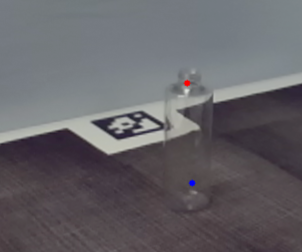}
                \centering{(a)}
            \end{minipage}
        \end{minipage}
        \hfill
        \begin{minipage}{0.16\linewidth}
            \begin{minipage}{1.0\linewidth}
                \includegraphics[width=1.0\linewidth]{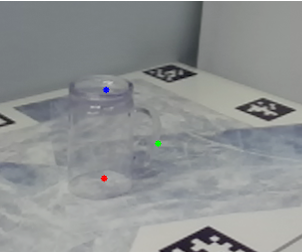}
            \end{minipage}
            \vfill
            \vspace{1mm}
            \begin{minipage}{1.0\linewidth}
                \includegraphics[width=1.0\linewidth]{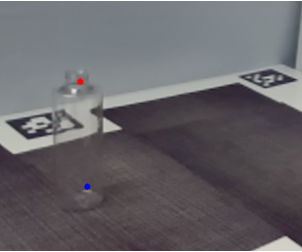}
                \centering{(b)}
            \end{minipage}
        \end{minipage}
        \hfill
        \begin{minipage}{0.16\linewidth}
            \begin{minipage}{1.0\linewidth}
                \includegraphics[width=1.0\linewidth]{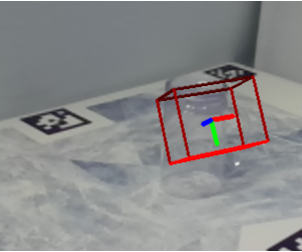}
            \end{minipage}
            \vfill
            \vspace{1mm}
            \begin{minipage}{1.0\linewidth}
                \includegraphics[width=1.0\linewidth]{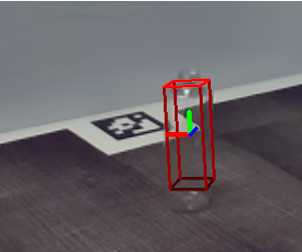}
                \centering{(c)}
            \end{minipage}
        \end{minipage}
        \hfill
        \begin{minipage}{0.16\linewidth}
            \begin{minipage}{1.0\linewidth}
                \includegraphics[width=1.0\linewidth]{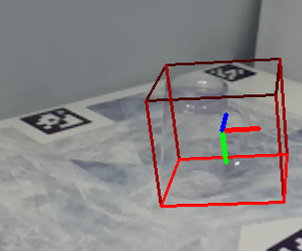}
            \end{minipage}
            \vfill
            \vspace{1mm}
            \begin{minipage}{1.0\linewidth}
                \includegraphics[width=1.0\linewidth]{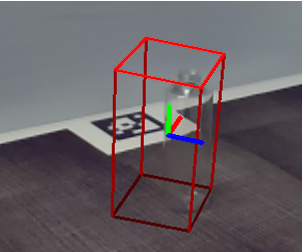}
                \centering{(d)}
            \end{minipage}
        \end{minipage}
        \hfill
        \begin{minipage}{0.16\linewidth}
            \begin{minipage}{1.0\linewidth}
                \includegraphics[width=1.0\linewidth]{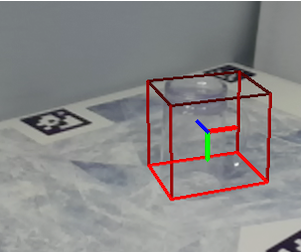}
            \end{minipage}
            \vfill
            \vspace{1mm}
            \begin{minipage}{1.0\linewidth}
                \includegraphics[width=1.0\linewidth]{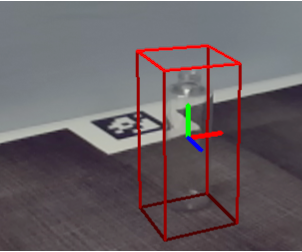}
                \centering{(e)}
            \end{minipage}
        \end{minipage}
        \hfill
        \begin{minipage}{0.16\linewidth}
            \begin{minipage}{1.0\linewidth}
                \includegraphics[width=1.0\linewidth]{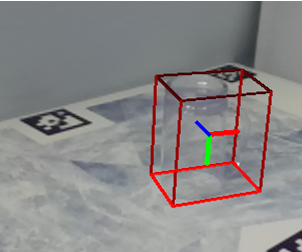}
            \end{minipage}
            \vfill
            \vspace{1mm}
            \begin{minipage}{1.0\linewidth}
                \includegraphics[width=1.0\linewidth]{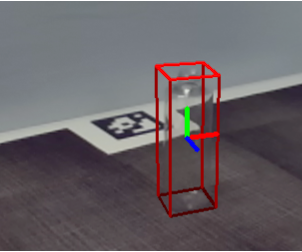}
                \centering{(f)}
            \end{minipage}
        \end{minipage}\vspace{-1mm}
    \caption{\textbf{Qualitative comparison with KeyPose.} (a) Predicted left-view keypoints. (b) Predicted right-view keypoints. (c) Pose results computed from key points. (d) Pose results w/o the proposed decoupled estimation scheme. (e) Pose results of the proposed method. (f) Ground-truth.}\vspace{-6mm}
    \end{minipage}
    \label{fig:qualitative_keypoint}
\end{figure*}

In addition, StereoPose also outperforms KeyPose, a key-point-based method for object pose estimation. As shown in Fig.~\ref{fig:qualitative_keypoint}, although the 2D positions of the predicted object key points on stereo images are visually accurate, the recovered object pose still exhibits a relatively large error in both rotation and translation. It is because the sparse object key points could provide only very limited correspondence information between the object frame and the camera frame. As a result, the recovered object pose is sensitive to the position error of predicted object key points. In contrast, the proposed StereoPose uses dense correspondences by predicting the NOCS map for robust object pose estimation. Moreover, the introduction of the back-view NOCS map, parallax-aware stereo feature fusion, and epipolar geometry loss function further improve the quality of the predicted NOCS map for a high pose accuracy.

\subsection{Ablation Studies}\label{subsec:ablation_study}
\begin{figure}
    \centering
    \begin{minipage}{0.93\linewidth}
        \begin{minipage}{1.0\linewidth}
            \begin{minipage}{0.235\linewidth}
                \includegraphics[width=1.0\linewidth]{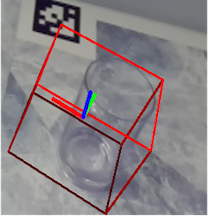}
            \end{minipage}
            \hfill
            \begin{minipage}{0.235\linewidth}
                \includegraphics[width=1.0\linewidth]{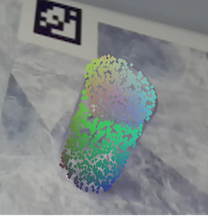}
            \end{minipage}
            \hfill
            \begin{minipage}{0.235\linewidth}
                \includegraphics[width=1.0\linewidth]{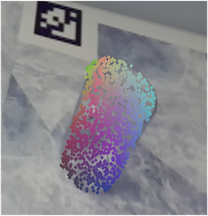}
            \end{minipage}
            \hfill
            \begin{minipage}{0.235\linewidth}
                \includegraphics[width=1.0\linewidth]{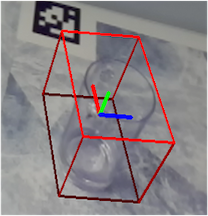}
            \end{minipage}
        \end{minipage}
        \vfill
        \vspace{1mm}
        \begin{minipage}{1.0\linewidth}
            \begin{minipage}{0.235\linewidth}
                \includegraphics[width=1.0\linewidth]{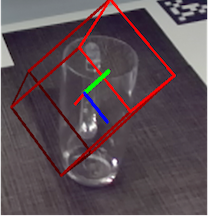}
                \centering{(a)}
            \end{minipage}
            \hfill
            \begin{minipage}{0.235\linewidth}
                \includegraphics[width=1.0\linewidth]{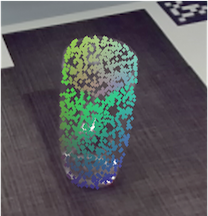}
                \centering{(b)}
            \end{minipage}
            \hfill
            \begin{minipage}{0.235\linewidth}
                \includegraphics[width=1.0\linewidth]{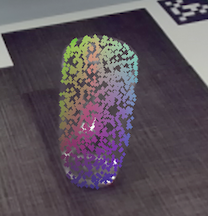}
                \centering{(c)}
            \end{minipage}
            \hfill
            \begin{minipage}{0.235\linewidth}
                \includegraphics[width=1.0\linewidth]{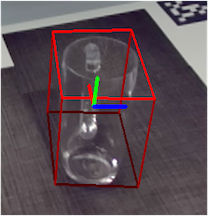}
                \centering{(d)}
            \end{minipage}
        \end{minipage}\vspace{-2mm}
    \caption{\textbf{Effect of the proposed back-view NOCS map.} (a) Results w/o back-view NOCS map. (b) Predicted front-view NOCS map. (c) Predicted back-view NOCS map. (d) Results w/ back-view NOCS map.}
    \end{minipage}\vspace{-7mm}
    \label{fig:qualitative_twoside}
\end{figure}

\textbf{Back-view NOCS Map.} We remove the back-view NOCS branch from the framework and evaluate the network on the same testing dataset. Tab.~\ref{table:ablation} presents the experiment results. Without the back-view NOCS map, all metrics suffer a severe drop. As shown in Fig.~\ref{fig:qualitative_twoside}, when the handle is on the back of the mug, the model without using the back-view NOCS map cannot localize the handle accurately and produces a large rotation error. In contrast, using the back-view NOCS map could leverage this informative structure cue on the back of the object and obtain a very accurate pose estimation result even in this challenging scenario. These results demonstrate the effectiveness of our proposed back-view NOCS map for transparent object pose estimation.

\textbf{Stereo Feature Fusion.} We remove the stereo feature fusion module from the framework and evaluate the network on the same testing dataset. Note that we still use both left and right images for network training and only remove the cross-view feature interaction module. Therefore, networks with and without stereo feature fusion are trained on the same number of images. As shown in Tab.~\ref{table:ablation}, the pose accuracy of the network without the stereo feature fusion module is significantly inferior to the result of our complete model. These results demonstrate the effectiveness of our proposed stereo fusion module in integrating the cross-view features for more accurate object pose estimation.

\begin{figure}
    \centering
    \begin{minipage}{0.93\linewidth}
        \begin{minipage}{1.0\linewidth}
            \includegraphics[width=1.0\linewidth]{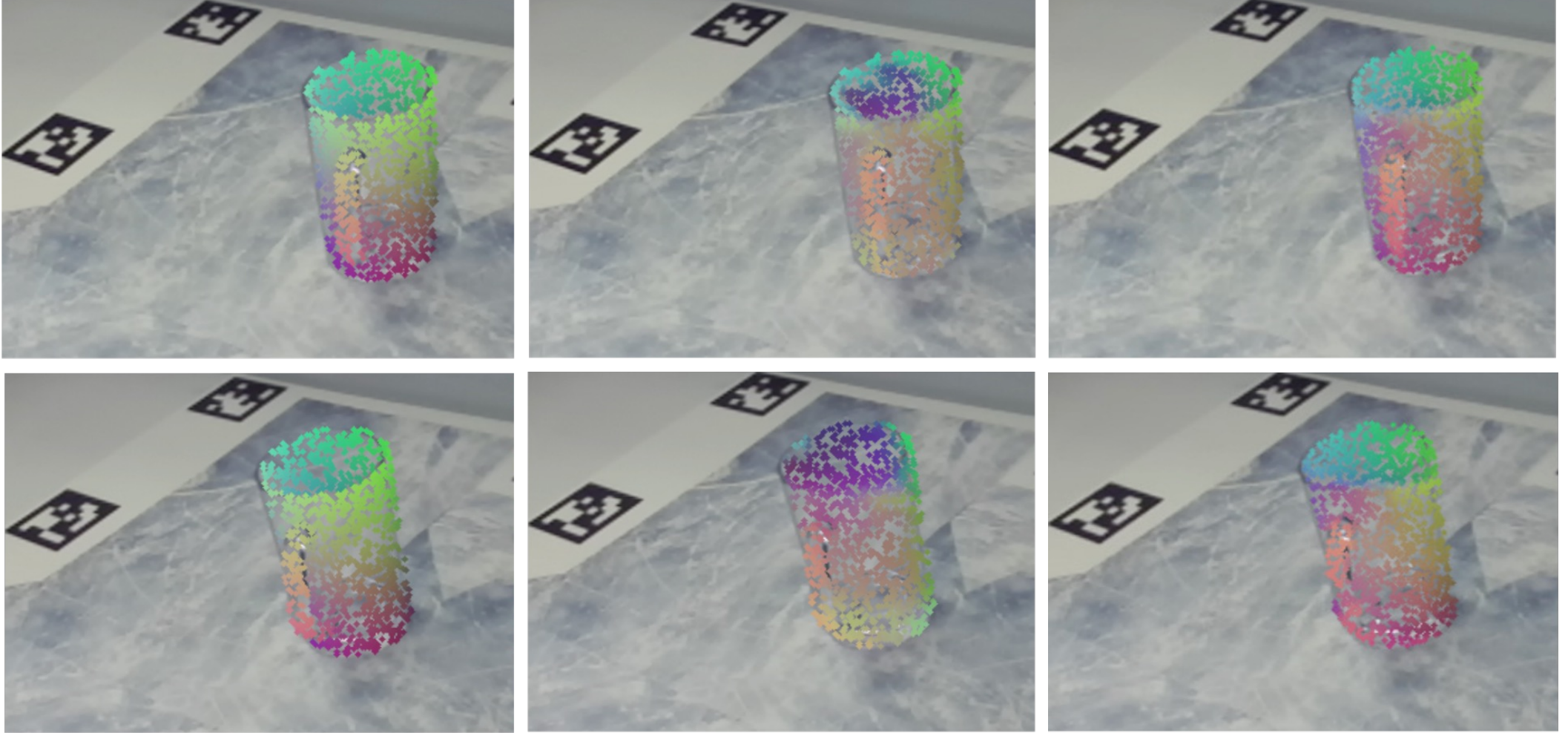}
        \end{minipage}
        \vfill
        \vspace{1mm}
        \begin{minipage}{1.0\linewidth}
            \begin{minipage}{0.32\linewidth}
                \centering{(a)}
            \end{minipage}
            \hfill
            \begin{minipage}{0.32\linewidth}
                \centering{(b)}
            \end{minipage}
            \hfill
            \begin{minipage}{0.32\linewidth}
                \centering{(c)}
            \end{minipage}
        \end{minipage}\vspace{-7mm}
        \caption{\textbf{Qualitative comparison of stereo NOCS maps predicted by models w/ and w/o the proposed epipolar loss.} (a) Ground-truth NOCS maps. (b) NOCS maps w/o $l_{ep}$. (c) NOCS maps w/ $l_{ep}$. The top row is for the left view and the bottom row is for the right view.}
    \end{minipage}\vspace{-5mm}
    \label{fig:qualitative_eploss}
\end{figure}

\textbf{Epipolar Geometry Loss.} We remove $l_{ep}$ when training the network and evaluate the model on the same testing dataset. Tab.~\ref{table:ablation} concludes the experiment results. Removing the epipolar loss hurts the pose accuracy in terms of nearly all evaluation metrics. Fig.~\ref{fig:qualitative_eploss} further presents the qualitative comparison on the NOCS map. We can find that the epipolar loss acting as a regularization term is effective in improving the quality of the predicted NOCS map. On the one hand, it improves the consistency between the left-view NOCS and the right-view NOCS. On the other hand, the epipolar loss also helps to produce a more accurate NOCS map when compared with the ground-truth NOCS map. These results demonstrate the effectiveness of the proposed epipolar loss for the stereo object pose estimation framework.\vspace{-3mm}

\begin{table}[thb!]
    \centering
    \caption{Ablation Study on Category of Mug.'A': Back-view NOCS map. 'B': Parallax attention fusion. 'C': Epipolar loss.}\vspace{-2mm}
    \label{table:ablation}
    \resizebox{\linewidth}{!}{%
    \renewcommand{\arraystretch}{0.9}
        \begin{tabular}{c|ccc|cccc}
            \hline
            {} & {A} & {B} & {C} & $3D_{25}$ & $3D_{50}$ & $10^{\circ}5cm$ & $10^{\circ}10cm$\\ \hline
            1 & - & - & - & 88.1 & 46.5 & 16.8 & 25.4\\
            2 & \checkmark & - & - & 93.3 & 55.4 & 19.8 & 29.3 \\
            3 & - & \checkmark & - & 89.9 & 48.5 & 18.7 & 27.9 \\
            4 & - & - & \checkmark & 96.4 & 63.2 & 21.3 & 24.2 \\
            5 & \checkmark & \checkmark & - & 95.5 & 75.2 & 26.8 & 31.0\\
            6 & \checkmark & - & \checkmark & 93.6 & 57.7 & 21.6 & 28.8 \\
            7 & - & \checkmark & \checkmark & 93.0 & 57.2 & 21.1 & 31.7 \\
            8 & \checkmark & \checkmark & \checkmark & \textbf{97.9} & \textbf{77.4} & \textbf{34.4} & \textbf{38.2}\\\hline
        \end{tabular}
    }\vspace{-7mm}
\end{table}
\section{Conclusion}\vspace{-2mm}
In this work, we propose StereoPose, a novel stereo image framework for category-level transparent object pose estimation.
StereoPose decouples category-level pose estimation into object size estimation, initial pose estimation, and pose refinement, achieving a robust and accurate category-level pose estimation for transparent objects with pure stereo images.
The back-view NOCS map is defined for transparent objects, which excavates the structure information on the back of the transparent object for more accurate pose estimation.
% We define the back-view NOCS map for transparent objects. The introduction of the back-view NOCS map reduces the ambiguity of predicting NOCS maps from the aliased image. In addition, it excavates the structure information on the back of the transparent object for more accurate pose estimation.
To further improve the performance of the stereo framework, we propose a parallax attention module for stereo feature fusion. An epipolar geometry loss is put forward to improve the consistency of stereo-view NOCS maps. 
Extensive experiments on the public TOD dataset demonstrate the superiority of the proposed StereoPose framework for category-level 6D transparent object pose estimation.

\bibliographystyle{IEEEtran}
\bibliography{ref}

\begin{thebibliography}{10}
\providecommand{\url}[1]{#1}
\csname url@rmstyle\endcsname
\providecommand{\newblock}{\relax}
\providecommand{\bibinfo}[2]{#2}
\providecommand\BIBentrySTDinterwordspacing{\spaceskip=0pt\relax}
\providecommand\BIBentryALTinterwordstretchfactor{4}
\providecommand\BIBentryALTinterwordspacing{\spaceskip=\fontdimen2\font plus
\BIBentryALTinterwordstretchfactor\fontdimen3\font minus
  \fontdimen4\font\relax}
\providecommand\BIBforeignlanguage[2]{{%
\expandafter\ifx\csname l@#1\endcsname\relax
\typeout{** WARNING: IEEEtran.bst: No hyphenation pattern has been}%
\typeout{** loaded for the language `#1'. Using the pattern for}%
\typeout{** the default language instead.}%
\else
\language=\csname l@#1\endcsname
\fi
#2}}

\bibitem{wang2019normalized}
H.~Wang, S.~Sridhar, J.~Huang, J.~Valentin, S.~Song, and L.~J. Guibas,
  ``Normalized object coordinate space for category-level 6d object pose and
  size estimation,'' in \emph{Proceedings of the IEEE/CVF Conference on
  Computer Vision and Pattern Recognition}, 2019, pp. 2642--2651.

\bibitem{deng2022icaps}
X.~Deng, J.~Geng, T.~Bretl, Y.~Xiang, and D.~Fox, ``icaps: Iterative
  category-level object pose and shape estimation,'' \emph{IEEE Robotics and
  Automation Letters}, vol.~7, no.~2, pp. 1784--1791, 2022.

\bibitem{liu2022catre}
X.~Liu, G.~Wang, Y.~Li, and X.~Ji, ``Catre: Iterative point clouds alignment
  for category-level object pose refinement,'' \emph{arXiv preprint
  arXiv:2207.08082}, 2022.

\bibitem{irshad2022shapo}
M.~Z. Irshad, S.~Zakharov, R.~Ambrus, T.~Kollar, Z.~Kira, and A.~Gaidon,
  ``Shapo: Implicit representations for multi-object shape, appearance, and
  pose optimization,'' \emph{arXiv preprint arXiv:2207.13691}, 2022.

\bibitem{chen2021sgpa}
K.~Chen and Q.~Dou, ``Sgpa: Structure-guided prior adaptation for
  category-level 6d object pose estimation,'' in \emph{Proceedings of the
  IEEE/CVF International Conference on Computer Vision}, 2021, pp. 2773--2782.

\bibitem{sajjan2020clear}
S.~Sajjan, M.~Moore, M.~Pan, G.~Nagaraja, J.~Lee, A.~Zeng, and S.~Song, ``Clear
  grasp: 3d shape estimation of transparent objects for manipulation,'' in
  \emph{2020 IEEE International Conference on Robotics and Automation
  (ICRA)}.\hskip 1em plus 0.5em minus 0.4em\relax IEEE, 2020, pp. 3634--3642.

\bibitem{zhu2021rgb}
L.~Zhu, A.~Mousavian, Y.~Xiang, H.~Mazhar, J.~van Eenbergen, S.~Debnath, and
  D.~Fox, ``Rgb-d local implicit function for depth completion of transparent
  objects,'' in \emph{Proceedings of the IEEE/CVF Conference on Computer Vision
  and Pattern Recognition}, 2021, pp. 4649--4658.

\bibitem{xu2021seeing}
H.~Xu, Y.~R. Wang, S.~Eppel, A.~Aspuru-Guzik, F.~Shkurti, and A.~Garg, ``Seeing
  glass: Joint point-cloud and depth completion for transparent objects,'' in
  \emph{5th Annual Conference on Robot Learning}, 2021.

\bibitem{fang2022transcg}
H.~Fang, H.-S. Fang, S.~Xu, and C.~Lu, ``Transcg: A large-scale real-world
  dataset for transparent object depth completion and a grasping baseline,''
  \emph{IEEE Robotics and Automation Letters}, vol.~7, no.~3, pp. 7383--7390,
  2022.

\bibitem{peng2019pvnet}
S.~Peng, Y.~Liu, Q.~Huang, X.~Zhou, and H.~Bao, ``Pvnet: Pixel-wise voting
  network for 6dof pose estimation,'' in \emph{Proceedings of the IEEE/CVF
  Conference on Computer Vision and Pattern Recognition}, 2019, pp. 4561--4570.

\bibitem{chen2022clearpose}
X.~Chen, H.~Zhang, Z.~Yu, A.~Opipari, and O.~C. Jenkins, ``Clearpose:
  Large-scale transparent object dataset and benchmark,'' \emph{arXiv preprint
  arXiv:2203.03890}, 2022.

\bibitem{ichnowski2021dex}
J.~Ichnowski, Y.~Avigal, J.~Kerr, and K.~Goldberg, ``Dex-nerf: Using a neural
  radiance field to grasp transparent objects,'' in \emph{5th Annual Conference
  on Robot Learning}, 2021.

\bibitem{zhu2021transfusion}
Y.~Zhu, J.~Qiu, and B.~Ren, ``Transfusion: A novel slam method focused on
  transparent objects,'' in \emph{Proceedings of the IEEE/CVF International
  Conference on Computer Vision}, 2021, pp. 6019--6028.

\bibitem{phillips2016seeing}
C.~J. Phillips, M.~Lecce, and K.~Daniilidis, ``Seeing glassware: from edge
  detection to pose estimation and shape recovery.'' in \emph{Robotics: Science
  and Systems}, vol.~3.\hskip 1em plus 0.5em minus 0.4em\relax Michigan, USA,
  2016, p.~3.

\bibitem{lysenkov2013recognition}
I.~Lysenkov, V.~Eruhimov, and G.~Bradski, ``Recognition and pose estimation of
  rigid transparent objects with a kinect sensor,'' \emph{Robotics}, vol. 273,
  no. 273-280, p.~2, 2013.

\bibitem{lysenkov2013pose}
I.~Lysenkov and V.~Rabaud, ``Pose estimation of rigid transparent objects in
  transparent clutter,'' in \emph{2013 IEEE International Conference on
  Robotics and Automation}.\hskip 1em plus 0.5em minus 0.4em\relax IEEE, 2013,
  pp. 162--169.

\bibitem{tang2021depthgrasp}
Y.~Tang, J.~Chen, Z.~Yang, Z.~Lin, Q.~Li, and W.~Liu, ``Depthgrasp: Depth
  completion of transparent objects using self-attentive adversarial network
  with spectral residual for grasping,'' in \emph{2021 IEEE/RSJ International
  Conference on Intelligent Robots and Systems (IROS)}.\hskip 1em plus 0.5em
  minus 0.4em\relax IEEE, 2021, pp. 5710--5716.

\bibitem{dai2022domain}
Q.~Dai, J.~Zhang, Q.~Li, T.~Wu, H.~Dong, Z.~Liu, P.~Tan, and H.~Wang, ``Domain
  randomization-enhanced depth simulation and restoration for perceiving and
  grasping specular and transparent objects,'' \emph{arXiv preprint
  arXiv:2208.03792}, 2022.

\bibitem{jiang2022a4t}
J.~Jiang, G.~Cao, T.-T. Do, and S.~Luo, ``A4t: Hierarchical affordance
  detection for transparent objects depth reconstruction and manipulation,''
  \emph{IEEE Robotics and Automation Letters}, vol.~7, no.~4, pp. 9826--9833,
  2022.

\bibitem{xu20206dof}
C.~Xu, J.~Chen, M.~Yao, J.~Zhou, L.~Zhang, and Y.~Liu, ``6dof pose estimation
  of transparent object from a single rgb-d image,'' \emph{Sensors}, vol.~20,
  no.~23, p. 6790, 2020.

\bibitem{chen2020learning}
D.~Chen, J.~Li, Z.~Wang, and K.~Xu, ``Learning canonical shape space for
  category-level 6d object pose and size estimation,'' in \emph{Proceedings of
  the IEEE/CVF conference on computer vision and pattern recognition}, 2020,
  pp. 11\,973--11\,982.

\bibitem{li2021leveraging}
X.~Li, Y.~Weng, L.~Yi, L.~J. Guibas, A.~Abbott, S.~Song, and H.~Wang,
  ``Leveraging se (3) equivariance for self-supervised category-level object
  pose estimation from point clouds,'' \emph{Advances in Neural Information
  Processing Systems}, vol.~34, pp. 15\,370--15\,381, 2021.

\bibitem{sahin2018category}
C.~Sahin and T.-K. Kim, ``Category-level 6d object pose recovery in depth
  images,'' in \emph{Proceedings of the European Conference on Computer Vision
  (ECCV) Workshops}, 2018, pp. 0--0.

\bibitem{sahin2019instance}
C.~Sahin, G.~Garcia-Hernando, J.~Sock, and T.-K. Kim, ``Instance-and
  category-level 6d object pose estimation,'' in \emph{RGB-D Image Analysis and
  Processing}.\hskip 1em plus 0.5em minus 0.4em\relax Springer, 2019, pp.
  243--265.

\bibitem{tian2020shape}
M.~Tian, M.~H. Ang, and G.~H. Lee, ``Shape prior deformation for categorical 6d
  object pose and size estimation,'' in \emph{European Conference on Computer
  Vision}.\hskip 1em plus 0.5em minus 0.4em\relax Springer, 2020, pp. 530--546.

\bibitem{wang2021category}
J.~Wang, K.~Chen, and Q.~Dou, ``Category-level 6d object pose estimation via
  cascaded relation and recurrent reconstruction networks,'' in \emph{2021
  IEEE/RSJ International Conference on Intelligent Robots and Systems
  (IROS)}.\hskip 1em plus 0.5em minus 0.4em\relax IEEE, 2021, pp. 4807--4814.

\bibitem{lin2022category}
J.~Lin, Z.~Wei, C.~Ding, and K.~Jia, ``Category-level 6d object pose and size
  estimation using self-supervised deep prior deformation networks,''
  \emph{arXiv preprint arXiv:2207.05444}, 2022.

\bibitem{irshad2022centersnap}
M.~Z. Irshad, T.~Kollar, M.~Laskey, K.~Stone, and Z.~Kira, ``Centersnap:
  Single-shot multi-object 3d shape reconstruction and categorical 6d pose and
  size estimation,'' \emph{arXiv preprint arXiv:2203.01929}, 2022.

\bibitem{di2022gpv}
Y.~Di, R.~Zhang, Z.~Lou, F.~Manhardt, X.~Ji, N.~Navab, and F.~Tombari,
  ``Gpv-pose: Category-level object pose estimation via geometry-guided
  point-wise voting,'' in \emph{Proceedings of the IEEE/CVF Conference on
  Computer Vision and Pattern Recognition}, 2022, pp. 6781--6791.

\bibitem{lin2022sar}
H.~Lin, Z.~Liu, C.~Cheang, Y.~Fu, G.~Guo, and X.~Xue, ``Sar-net: Shape
  alignment and recovery network for category-level 6d object pose and size
  estimation,'' in \emph{Proceedings of the IEEE/CVF Conference on Computer
  Vision and Pattern Recognition}, 2022, pp. 6707--6717.

\bibitem{zhang2022rbp}
R.~Zhang, Y.~Di, Z.~Lou, F.~Manhardt, N.~Navab, F.~Tombari, and X.~Ji,
  ``Rbp-pose: Residual bounding box projection for category-level pose
  estimation,'' \emph{arXiv preprint arXiv:2208.00237}, 2022.

\bibitem{lepetit2009epnp}
V.~Lepetit, F.~Moreno-Noguer, and P.~Fua, ``Epnp: An accurate o (n) solution to
  the pnp problem,'' \emph{International journal of computer vision}, vol.~81,
  no.~2, pp. 155--166, 2009.

\bibitem{fischler1981random}
M.~A. Fischler and R.~C. Bolles, ``Random sample consensus: a paradigm for
  model fitting with applications to image analysis and automated
  cartography,'' \emph{Communications of the ACM}, vol.~24, no.~6, pp.
  381--395, 1981.

\bibitem{haugaard2022surfemb}
R.~L. Haugaard and A.~G. Buch, ``Surfemb: Dense and continuous correspondence
  distributions for object pose estimation with learnt surface embeddings,'' in
  \emph{Proceedings of the IEEE/CVF Conference on Computer Vision and Pattern
  Recognition}, 2022, pp. 6749--6758.

\bibitem{yin2020disentangled}
M.~Yin, Z.~Yao, Y.~Cao, X.~Li, Z.~Zhang, S.~Lin, and H.~Hu, ``Disentangled
  non-local neural networks,'' in \emph{European Conference on Computer
  Vision}.\hskip 1em plus 0.5em minus 0.4em\relax Springer, 2020, pp. 191--207.

\bibitem{wang2021symmetric}
Y.~Wang, X.~Ying, L.~Wang, J.~Yang, W.~An, and Y.~Guo, ``Symmetric parallax
  attention for stereo image super-resolution,'' in \emph{Proceedings of the
  IEEE/CVF Conference on Computer Vision and Pattern Recognition}, 2021, pp.
  766--775.

\bibitem{hartley2003multiple}
R.~Hartley and A.~Zisserman, \emph{Multiple view geometry in computer
  vision}.\hskip 1em plus 0.5em minus 0.4em\relax Cambridge university press,
  2003.

\bibitem{liu2020keypose}
X.~Liu, R.~Jonschkowski, A.~Angelova, and K.~Konolige, ``Keypose: Multi-view 3d
  labeling and keypoint estimation for transparent objects,'' in
  \emph{Proceedings of the IEEE/CVF conference on computer vision and pattern
  recognition}, 2020, pp. 11\,602--11\,610.

\bibitem{you2022cppf}
Y.~You, R.~Shi, W.~Wang, and C.~Lu, ``Cppf: Towards robust category-level 9d
  pose estimation in the wild,'' in \emph{Proceedings of the IEEE/CVF
  Conference on Computer Vision and Pattern Recognition}, 2022, pp. 6866--6875.

\bibitem{he2016deep}
K.~He, X.~Zhang, S.~Ren, and J.~Sun, ``Deep residual learning for image
  recognition,'' in \emph{Proceedings of the IEEE conference on computer vision
  and pattern recognition}, 2016, pp. 770--778.

\bibitem{zhao2017pyramid}
H.~Zhao, J.~Shi, X.~Qi, X.~Wang, and J.~Jia, ``Pyramid scene parsing network,''
  in \emph{Proceedings of the IEEE conference on computer vision and pattern
  recognition}, 2017, pp. 2881--2890.

\bibitem{xie2020segmenting}
E.~Xie, W.~Wang, W.~Wang, M.~Ding, C.~Shen, and P.~Luo, ``Segmenting
  transparent objects in the wild,'' in \emph{European conference on computer
  vision}.\hskip 1em plus 0.5em minus 0.4em\relax Springer, 2020, pp. 696--711.

\end{thebibliography}
%%%%%%%%%%%%%%%%%%%%%%%%%%%%%%%%%%%%%%%%%%%%%%%%%%%%%%%%%%%%%%%%%%%%%%%%%%%%%%%%
\end{document}